\documentclass{article}
\usepackage{ijcai26}

\usepackage{times}
\usepackage{soul}
\usepackage{url}
\usepackage[hidelinks]{hyperref}
\usepackage[utf8]{inputenc}
\usepackage[small]{caption}
\usepackage{graphicx}
\usepackage{amsmath}
\usepackage{amsthm}
\usepackage{booktabs}
\usepackage{algorithm}
\usepackage{algorithmic}
\usepackage[switch]{lineno}
\usepackage{balance}
\usepackage{booktabs}
\usepackage{subfigure}
\usepackage{multicol,lipsum}
\usepackage{multirow}
\usepackage{amssymb}
\usepackage{bbding}
\usepackage{pifont}
\usepackage{wasysym}
\usepackage{utfsym}
\usepackage{fontawesome}
\usepackage{multirow}
\newtheorem{definition}{Definition}

\title{\textsc{DGCPath}: Distribution-Aware Generative Contrastive Framework for Self-supervised Path Representation Learning -- Extended Version}

\author{
Sean Bin Yang$^1$
\and
Hao Miao$^2$ \and
Zongyi Xu$^{3}$\and
Jilin Hu$^4$\and \\
Xiangmeng Wang$^2$\and 
Hua Lu$^1$\and
Bin Yang$^{1,4}$\and
Christian S. Jensen$^1$\\
\affiliations
$^1$Aalborg University, Denmark\\
$^2$The Hong Kong Polytechnic University, Hong Kong \\
$^3$Chongqing Univeristy of Posts and Telecommunications, China\\
$^4$East China Normal University, China\\
\emails
\{seany, luhua, csj\}@cs.aau.dk,
\{hao.miao, xiangmengpoly.wang\}@polyu.edu.hk, \\
xuzy@cqupt.edu.cn,
\{jlhu, byang\}@dase.ecnu.edu.cn
}

\vspace{-10pt}
\begin{document}

\maketitle

\begin{abstract}
    Due to proliferation of vehicle trajectory data from advanced sensing technologies, path representation learning has become a pivotal task in intelligent transportation systems. Although existing self-supervised approaches work to some extent, their dependence on deterministic contrastive learning paradigms and handcrafted view augmentation strategies inherently restricts cross-scenario generalization capabilities. To address these limitations, we present \textsc{DGCPath} – an innovative \textbf{D}istribution-aware \textbf{G}enerative \textbf{C}ontrastive learning framework for \textbf{P}ath representation. This architecture establishes a synergistic connection between generative modeling and distributional contrastive learning, enabling the acquisition of robust and transferable feature embeddings. Specifically, our framework incorporates: (1) a diffusion-based view generator that autonomously produces semantically coherent yet diversified trajectory views from Gaussian noise distributions; (2) a variational contrastive mechanism enforcing latent feature alignment at the distribution level, transcending conventional instance-wise consistency; and (3) a novel generative cross-supervision module that reinforces view-level consistency through cross-view reconstruction learning. Comprehensive evaluations on three real-world trajectory datasets demonstrate \textsc{DGCPath}'s superior performance over state-of-the-art baselines in two distinct downstream tasks, validating its enhanced generalization capacity and representation effectiveness.
    
    	\textbf{This is an extended version of '' \textsc{DGCPath}: Distribution-Aware Generative Contrastive Framework for Self-supervised Path Representation Learning'' \cite{YangDGC026}, to appear in IJCAI 2026.}
\end{abstract}

\section{Introduction}
Increasingly large volumes of vehicle path data are being accumulated that hold the potential to enable an expanding range of analyses in intelligent transportation systems \cite{DBLP:conf/icde/JensenYGHT24,DBLP:conf/aaai/00220ZDL25,DBLP:conf/kdd/YangHGYJ23,DBLP:conf/ijcai/YangGHT021,DBLP:conf/aaai/HanWWYLLW25,DBLP:journals/tits/LiYZLCP15,li2018extended,li2016sliding,li2017new,DBLP:journals/corr/abs-2411-18428,DBLP:journals/corr/abs-2411-18428}.
Representation learning for such data is a key enabler of such analyses, as it represents raw paths by compact and informative low-dimensional vectors \cite{DBLP:conf/kdd/YangHGYJ23,DBLP:conf/ijcai/YangGHT021,DBLP:conf/icde/YangGHYTJ22,TFM,DBLP:journals/corr/abs-2106-09373,DBLP:journals/corr/abs-2203-16110}.
These representations facilitate intelligent transportation applications, such as traffic analysis \cite{DBLP:conf/aaai/00220ZDL25,DBLP:journals/corr/abs-2411-14768,DBLP:conf/icde/JiangPRJLW23,DBLP:journals/vldb/WuWYZGQHSJ24,DBLP:journals/pacmmod/LinWHGYLJ23,DBLP:conf/aaai/Mao00CXWXLW25,DBLP:conf/kdd/YuWSQZ0X24,DBLP:journals/pvldb/PanWZY0CGWTDZYZ23}, path recommendation \cite{DBLP:journals/tkde/YangGY22,MM-Path}, and traffic prediction \cite{Path-LLM,DBLP:conf/kdd/YangHGYJ23,DBLP:journals/pvldb/FangPCDG21}.

\begin{figure}[!t]
	\centering
	\includegraphics[scale=0.73]{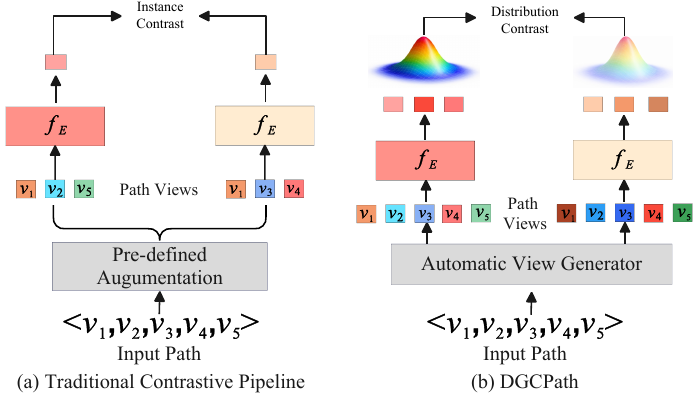}
	\caption{Traditional Contrastive Pipeline vs. \textsc{DGCPath}.}
	\label{fig:example1}
	\vspace{-10pt}
\end{figure}

Thus, substantial efforts have been devoted to learning generic path representations capable of supporting a wide range of downstream applications \cite{DBLP:conf/kdd/YangHGYJ23,Path-LLM,DBLP:conf/ijcai/YangGHT021,MM-Path,DBLP:conf/icde/YangGHYTJ22}.
However, the reliance of existing proposals on deterministic contrastive loss functions is a core limitation. As illustrated in Figure~\ref{fig:example1}(a), most existing proposals adopt a traditional contrastive learning pipeline that contrasts two deterministic path samples (i.e., Instance Contrast) to learn generic path representations. For example, \textsc{PIM} \cite{DBLP:conf/ijcai/YangGHT021} employs an InfoNCE-based loss to align deterministic positive samples and separate negative ones. 
In contrast, as shown in Figure \ref{fig:example1}(b), the contrastive formulation in Figure \ref{fig:example1}(a) is a special case of the more general distribution-based contrastive training. The former's rigid, deterministic regime limits the expressiveness of learned representations and reduces their ability to generalize in dynamic, real-world scenarios. 

A further key limitation of existing proposals is their use of pre-defined view augmentation strategies. For instance, methods like \textsc{START} \cite{DBLP:conf/icde/JiangPRJLW23} generate different path views through node dropping (e.g., as shown in Figure \ref{fig:example1}(a)) to obtain augmented path views and optimize them using deterministic contrastive objectives. Although this general approach enables strong empirical performance, it relies heavily on the use of carefully chosen augmentations, often requiring extensive trial-and-error tuning. This static and inflexible design constrains the adaptability of learned representations, particularly when transferred across difference tasks. 

To overcome these limitations, we introduce \textsc{DGCPath}, a novel \textbf{D}istribution-aware \textbf{G}enerative \textbf{C}ontrastive learning framework for \textbf{P}ath representation learning. Rather than enforcing consistency on deterministic feature vectors, \textsc{DGCPath} models each path representation as a latent probability distribution, enabling the learning of continuous, smooth, and uncertainty-aware embeddings.
By aligning distributions rather than fixed instances, \textsc{DGCPath} relaxes the rigid constraints inherent to deterministic proposals and mitigates overfitting to specific samples \cite{DBLP:journals/corr/KingmaW13}. The probabilistic formulation and the resulting enhanced flexibility of the encoder improves the generalization capabilities of the learned representations markedly across different tasks.

\textsc{DGCPath} consists of three main components: a path encoder, a path view generator, and a generative variational–contrastive component. First, the path encoder maps each path into a latent representation by jointly capturing its spatial structure and temporal dynamics. Second, we introduce a diffusion-based path view generator that automatically synthesizes diverse yet semantically consistent path views from random Gaussian noise. This generative mechanism facilitates the construction of informative contrastive pairs without relying on manually designed augmentation heuristics. 

Next, we leverage contrastive learning to constrain the distribution of latent features such that samples originating from the same path are encouraged to share similar distributions, while those from different paths are separated in the latent space. Specifically, we unify generative modeling with contrastive learning by proposing a generative variational contrastive model that integrates the strengths of both paradigms: capturing path-invariant characteristics via generative learning, and emphasizing inter-path discriminability through contrastive objectives.
We also propose a generative cross-supervision module that replaces traditional self-supervision with a cross-view alignment mechanism. This design promotes consistency in the learned feature distributions across different views of the same path, thereby enhancing the robustness and semantic fidelity of the representations.

The main contributions are as follows:
\begin{itemize}
	\item We propose \textsc{DGCPath} that integrates generative modeling and distribution-based contrastive learning to learn robust and generalizable path representations.
	\item We propose a diffusion-based path view generator that automatically produces semantically consistent yet diverse path views from noise, eliminating the need for handcrafted augmentation.
	\item We propose a generative cross-supervision module and distributional contrastive loss to enforce consistency among views of the same path while promoting separation across different paths at the distribution level.
	\item We report on extensive experiments on three real-world datasets and considering two downstream tasks, showing that \textsc{DGCPath} can consistently outperform state-of-the-art methods in terms of both representation quality and downstream performance.
	
\end{itemize}

\section{Related Work}

\paragraph{Path Representation Learning} Path representation learning has been studied widely due to its key role in facilitating intelligent transportation system analyses \cite{MM-Path,Path-LLM,DBLP:conf/ijcai/YangGHT021,DBLP:conf/icde/YangGHYTJ22,DBLP:conf/kdd/YangHGYJ23,DBLP:conf/aaai/00220ZDL25,DBLP:journals/pvldb/ZhouSCJK24,DBLP:journals/corr/abs-2411-14768,DBLP:conf/icde/JiangPRJLW23}. Most existing methods focus on single-modality data (i.e., path sequences). For instance, \textsc{PIM} \cite{DBLP:conf/ijcai/YangGHT021} utilizes curriculum-based negative sampling for contrastive learning on graph-based path embedding, and \textsc{TPR} \cite{DBLP:conf/icde/YangGHYTJ22} enhances temporal path representations through temporally-informed view construction, while \textsc{LightPath} \cite{DBLP:conf/kdd/YangHGYJ23} introduces a lightweight and scalable framework using mask-ratio-based path view generation.
From a multimodal perspective, \textsc{MM-Path} \cite{MM-Path} incorporates path–image pairs and learns general representations via path–image contrastive learning, whereas \textsc{Path-LLM} \cite{Path-LLM} leverages path textual descriptions and large language models (LLMs) to enrich path semantics. We do not include these methods as baselines, as our study focuses exclusively on the path modality.
However, two key limitations remain. First, except for multimodal approaches, existing methods generally rely on manually designed or fixed path view generation strategies. Second, both generative and contrastive objectives are typically deterministic, which can limit the diversity and generalizability of learned path representations.

\section{Preliminaries}
\begin{definition}
	\textbf{(Road Network) } A road network is modeled as a directed graph $\mathit{G} = (\mathit{V}, \mathit{E})$, where $\mathit{V}$ denotes a set of $\mathit{N}$ nodes representing  road intersections, and  $\mathit{E}$ denotes a set of directed $\mathit{M}$ edges representing road segments.
\end{definition}

\begin{definition}
	\textbf{(Path) }  A path $\mathit{p} = \langle v_1, v_2, v_3, \ldots, v_{|\mathit{p}|} \rangle$ is an ordered sequence of connected nodes. Consecutive nodes in the sequence are connected by directed edges, i.e., $(v_i, v_{i+1}) \in \mathit{E}$ for all $1 \leq i < |\mathit{p}|$, ensuring the topological continuity of the path.
\end{definition}

\begin{figure*}[!t]
	\centering
	\includegraphics[scale=1]{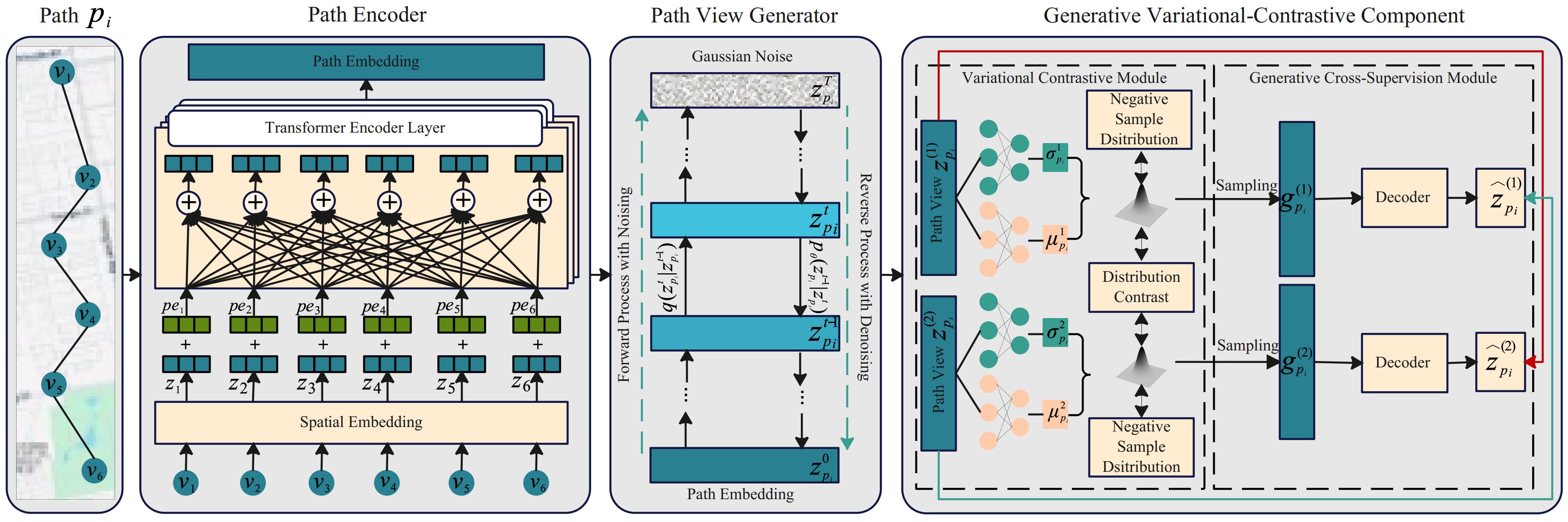}
	\caption{The \textsc{DGCPath} Framework.}
\label{fig:framework}
\vspace{-10pt}
\end{figure*}
\paragraph{Problem Statement} Given a road network $\mathit{G}$ and a set of paths $\mathit{P} = \{p_i\}_{i=1}^{\mathbb{N}}$, the objective is to learn a mapping function $\mathit{f}_{\theta}: \mathit{P} \rightarrow \mathbb{R}^{d}$, such that $\mathit{f}_\theta$ maps each path $p_i \in \mathit{P}$ to a $d$-dimensional representation vector in a latent space, with $\theta$ denoting the learnable parameters of the function.

\section{Methodology}
The \textsc{DGCPath} framework is illustrated in Figure \ref{fig:framework}. The core idea is to integrate the path encoder, the diffusion-based path view generator, and the generative variational contrastive component to enable more generic path represenation learning. 

\subsection{Path Encoder}

\paragraph{Spatial Embedding}The encoder aims to learn segment embeddings that preserve the topological structure of the road network, as such representations are fundamental to path modeling. Given a road network graph $\mathit{G} = (\mathit{V}, \mathit{E})$, we employ the \textsc{Node2vec} algorithm~\cite{DBLP:conf/kdd/GroverL16} to initialize spatial embeddings for each node. These embeddings capture the structural context of the network and serve as the basis for the subsequent path representation learning, which is formulated as follows:
\vspace{-5pt}
\begin{equation}
\mathit{z}_{\mathit{G}}=\mathit{Node2vec}(\mathit{G}(\mathit{V},\mathit{E})),
\end{equation}
\noindent
where $\mathit{z}_{\mathit{G}} \in \mathbb{R}^{d_n}$ is a set of node embeddings for road network ${\mathit{G}}$ and $d_n$ is the feature dimensionality.

\paragraph{Path Embedding} The path embedding module aims to learn path representations by aggregating consecutive road segment embeddings through spatial embedding. As shown in Figure \ref{fig:framework}, for a given path $p_i=\langle v_1,v_2,v_3,v_4,v_5,v_6\rangle$, the spatial embedding module first transforms the sequence of nodes in a
path to a sequence of node embeddings for the path, i.e., $\langle z_1,z_2,\ldots,z_6\rangle=\mathit{z}_{\mathit{G}}(\langle v_1,v_2,\ldots,v_t\rangle)$. Next, we employ the Transformer architecture to process the spatial embeddings (i.e., the $z_i$) and the positional embedding of the road segments within the path, computing the path embedding as follows:
\vspace{-5pt}
\begin{equation}
\begin{aligned}
	\mathit{h}_{p_i}&=  \langle \mathit{z}_{\mathit{G}}(v_1)+pe_1, \ldots, \mathit{z}_{\mathit{G}}(v_i)+pe_i \rangle, \\
	\mathit{z}_{p_i}&= \sigma\bigg( \sum \frac{\mathit{Transformer}(\mathit{h}_{p_i})}{|p_i|}\bigg),
\end{aligned}
\end{equation}
\noindent
where the $v_i$ are the nodes in the path, $\mathit{z}_{\mathit{G}}$ is the spatial embedding function, the $h_{p_i}$ captures the node embeddings after incorporating the position embedding $pe_i$, $|p_i|$ is the number of nodes in a path, and $z_{p_i}$ is the resulting path embedding. The function $\mathit{Transformer}(\cdot)$ refers to a Transformer architecture based on the self-attention mechanism. 

\subsection{Path View Generator}
To facilitate path encoder learning, we propose an automatic path view generation strategy based on a diffusion and denoising process applied to the initial path embedding.

\paragraph{Forward Process}Conditioned on the path embedding $\mathit{z}_{p_i}$ generated by the path encoder, we aim to learn the underlying distribution of path embeddings to enable automatic path view generation. This distribution is modeled using a generative diffusion process, where the values of the path embedding vector are diffused progressively and then denoised over discrete time steps. Specifically, given the initial path embedding $\mathit{z}_{p_i}$, we formulate the diffusion process as a Markov chain $\mathit{z}_{p_i}=(\mathit{z}_{p_i}^{0}, \mathit{z}_{p_i}^{1}, \ldots, \mathit{z}_{p_i}^{T})$, where $T$ denotes the total number of diffusion steps.
Going from $\mathit{z}_{p_i}^{0}$ to $\mathit{z}_{p_i}^{T}$ (i.e., the forward process with noising), we iteratively add small amounts of Gaussian noise, gradually corrupting the original path embedding until it becomes pure Gaussian noise. This process has an analogy in the image domain, where noise is typically applied independently across dimensions \cite{DBLP:journals/csur/YangZSHXZZCY24}. The corruption of path embeddings is governed by the following transition probabilities:
\vspace{-5pt}
\begin{equation}
\mathit{q}(\mathit{z}_{p_i}^{t}|\mathit{z}_{p_i}^{t-1}) :=\mathcal{N}\left(z_{p_i}^{t};\sqrt{1-\beta_t} z^t_{p_i},\beta_t I \right),
\end{equation}
\noindent
where $\beta_t=1-\alpha_t$ and $\bar{\alpha}_t=\prod_{s=1}^t \alpha_t$.

\paragraph{Reverse Process}The view generation process for the path embedding $\mathit{z}_{p_i}$ is formulated as a reverse denoising procedure, iteratively refining the embedding from $\mathit{z}_{p_i}^{T}$ back to $\mathit{z}_{p_i}^{0}$. Specifically, starting from the fully noised embedding $\mathit{z}_{p_i}^{T}$, we progressively remove the noise step by step to ultimately recover a sample from the original data distribution. This process is defined as follows:
\vspace{-5pt}
\begin{equation}
\mathit{p}_{\theta}(\mathit{z}_{p_i}^{t-1}|\mathit{z}_{p_i}^{t}) :=\mathcal{N}\left(\mathit{z}_{p_i}^{t-1};\mu_{\theta}(\mathit{z}_{p_i}^{t},t),\Sigma_{\theta}(\mathit{z}_{p_i}^{t},t)\right),
\end{equation}
\noindent
where $\Sigma_{\theta}$ is set to fixed, time-dependent constants to simplify the computation. To complete the reverse process, we incorporate a trainable rounding step $\mathit{p}_{\theta}(\mathit{z}_{p_i} \mid \mathit{z}_{p_i}^{0})$ that maps the denoised hidden states back to the final path view embedding. Specifically, we employ a learnable model $\mathit{f}_{\theta}(\mathit{z}_{p_i}^{t}, t)$ to parameterize this reverse mapping, effectively modeling $\mathit{p}_{\theta}(\mathit{z}_{p_i} \mid \mathit{z}_{p_i}^{0})$.

To train the diffusion model, we adopt a Kullback–Leibler (KL) divergence-based objective to enable more efficient optimization. This approach compares the reverse transition distribution $\mathit{p}_{\theta}(\mathit{z}_{p_i}^{t-1} \mid \mathit{z}_{p_i}^{t})$ with the corresponding posterior of the forward process, which leads to a simplified mean squared error (MSE) training loss, denoted as $\mathcal{L}_{\mathit{diff}}$:
\vspace{-5pt}
\begin{equation}
\mathcal{L}_{\mathit{diff }}\left(\mathit{z}_{p_i}^0\right)=\sum_{t=1}^T \underset{q\left(\mathit{z}_{p_i}^t \mid \mathit{z}_{p_i}^0 \right)}{\mathbb{E}}\left\|\mu_\theta\left(\mathit{z}_{p_i}^t , t \right)-\hat{\mu}\left(\mathit{z}_{p_i}^t , \mathit{z}_{p_i}^0 \right)\right\|^2,
\end{equation}
\noindent
where $\mu_\theta\left(\mathit{z}_{p_i}^t , t \right)$ is the predicted mean of $\mathit{p}_{\theta}(\mathit{z}_{p_i}^{t-1} \mid \mathit{z}_{p_i}^{t})$ computed by a neural network, and $\hat{\mu}\left(\mathit{z}_{p_i}^t , \mathit{z}_{p_i}^0 \right)$ denotes the mean of the posterior $\mathit{q}(\mathit{z}_{q_i}^{t-1}|\mathit{z}_{q_i}^{t},\mathit{z}_{q_i}^{0})$, which is tractable when conditioned on $\mathit{z}_{q_i}^{0}$.

\paragraph{Automatic View Generation} In contrastive learning, the diffusion model is employed as a path view generator to obtain multiple diverse yet semantically consistent representations of a given path embedding (i.e., $\mathit{z}_{p_i}^{(1)}$ and $\mathit{z}_{p_i}^{(2)}$). These generated views serve to improve the quality and robustness of contrastive representation learning.
Given a path embedding $\mathit{z}_{p_i}$, we begin by sampling a noise vector from a standard Gaussian distribution, denoted as $\tilde{\mathit{z}}_{p_i}^T \sim \mathcal{N}(0, \mathrm{I})$, where $\mathrm{I}$ is the identity matrix. Starting from this noisy input, we iteratively apply the reverse diffusion process to progressively denoise the representation and recover a refined embedding $\mathit{z}_{p_i}^{0}$.
In each reverse step, the process involves two key operations: 1) A rounding step that maps the intermediate state $\mathit{z}_{p_i}^{t}$ back into the path embedding space; 2) a replacement step in which certain components of the recovered embedding $\mathit{z}_{p_i}^{t-1}$ are substituted with corresponding segments from the original path embedding $\mathit{z}_{p_i}$ to retain critical semantic information.
This iterative procedure yields a single generated path view. To obtain a pair of views for contrastive learning, the process is performed twice with independently sampled Gaussian noise, resulting in $\mathit{z}_{p_i}^{(1)}$ and $\mathit{z}_{p_i}^{(2)}$ (see Figure~\ref{fig:framework}). While originating from the same input path embedding, the stochastic nature of the initial noise introduces meaningful variation, which is essential for learning robust and discriminative representations through contrastive objectives.

\subsection{Generative Variational Contrastive Module}
We further detail the design of our distribution-aware generative contrastive module, which explicitly aligns distributions of positive samples while maximizing divergence from negative ones in a generative latent space \cite{DBLP:journals/pami/WangTYWDG24}. 
\paragraph{Distributional Contrastive Learning}For each pair of path views, $\mathit{z}_{p_i}^{(1)}$ and $\mathit{z}_{p_i}^{(2)}$, generated by the automatic view generator, two independent fully connected layers, denoted as $\mathit{f}_{\mu}(\cdot)$ and $\mathit{f}_{\sigma}(\cdot)$, are employed to project the features into a high-dimensional invariant space. As a result, each sample yields two sets of parameters: ($\mu_{p_i}^{(1)}$, $\sigma_{p_i}^{(1)}$), ($\mu_{p_i}^{(2)}$, $\sigma_{p_i}^{(2)}$),  corresponding to the mean and standard deviation of each view. We assume that the latent features of the two path views follow Gaussian distributions parameterized by the respective mean and standard deviation vectors.
\vspace{-5pt}
\begin{equation}
\mathit{z}_{p_i}^{(1)}(\mathit{x}) \sim \mathcal{N}(\mathit{x}; \mu_{p_i}^{(1)},\sigma_{p_i}^{(1)}), \mathit{z}_{p_i}^{(2)}(\mathit{x}) \sim \mathcal{N}(\mathit{x}; \mu_{p_i}^{(2)},\sigma_{p_i}^{(2)})
\end{equation}

The KL divergence can be employed as a regularization term to enforce that the latent representations of all samples approximate a standard normal distribution, with mean $0$ and standard deviation $1$.
\begin{equation}
\mathit{D}_{\mathit{K L}}(\mathcal{N}(\mu, \sigma), \mathcal{N}(0,1))
\end{equation}

In generative tasks, encouraging the model to produce samples similar to the training set is often beneficial. However, in tasks such as path recommendation and traffic analysis, forcing all samples to follow a uniform distribution can weaken the discriminative capabilities of the feature extractor. To address this issue, our method aims to constrain samples from the same path (i.e., positive samples) to follow similar Gaussian distributions, while ensuring that samples from different paths (i.e., negative samples) are mapped to distributions that are as dissimilar as possible. 
Achieving this objective involves computing the distance between two distributions.

A natural choice is to use the KL divergence, commonly used in Variational Autoencoders (VAEs) and defined as follows:
\vspace{-5pt}
\begin{equation}
\mathit{D}_{\mathit{K L}}(	\mathit{z}_{p_i}^{(1)}(\mathit{x}) \| \mathit{z}_{p_i}^{(2)}(\mathit{x}))=\int 	\mathit{z}_{p_i}^{(1)}(\mathit{x}) \ln \frac{	\mathit{z}_{p_i}^{(1)}(\mathit{x})}{\mathit{z}_{p_i}^{(2)}(\mathit{x})} \mathit{d x}.
\end{equation}

However, the KL divergence is inherently asymmetric, meaning that it yields different values depending on the order of the input $\mathit{z}_{p_i}^{(1)}(\mathit{x}) $ and $\mathit{z}_{p_i}^{(2)}(\mathit{x}) $. In \textsc{VAE}s \cite{DBLP:journals/corr/KingmaW13}, this asymmetry is not problematic, as all latent representations are explicitly regularized toward a fixed standard normal distribution with zero mean and unit variance, yielding a unidirectional mapping. However, in contrastive learning, the goal is to estimate the similarity between distributions of two augmented path views derived from the same input path. Thus, the similarity measure must be order-invariant to ensure that the positive pairs are treated symmetrically, regardless of the order in which they occur as arguments.To achieve this, we adopt the Jensen–Shannon (JS) divergence, a symmetric variant of the KL divergence. The JS divergence computes the similarity between two distributions in an order-invariant manner, making it well-suited for contrastive learning, where symmetric comparison of augmented views is essential. Formally, the JS divergence between two distributions $\mathit{z}_{p_i}^{(1)}(\mathit{x}) $ and $\mathit{z}_{p_i}^{(2)}(\mathit{x}) $ is defined as follows: 
\vspace{-5pt}
\begin{equation}
	\small
\begin{aligned}
	&\mathit{D}_{\mathit{JS}}(\mathit{z}_{p_i}^{(1)}(\mathit{x}), \mathit{z}_{p_i}^{(2)}(\mathit{x}))
	= \frac{1}{2}\mathit{D}_{\mathit{KL}}\left(\mathit{z}_{p_i}^{(1)}(\mathit{x}) \| 
	\frac{\mathit{z}_{p_i}^{(1)}(\mathit{x})+\mathit{z}_{p_i}^{(2)}(\mathit{x})}{2}\right)
	\notag\\
	&\quad + \frac{1}{2}\mathit{D}_{\mathit{KL}}\left(\mathit{z}_{p_i}^{(2)}(\mathit{x}) \| 
	\frac{\mathit{z}_{p_i}^{(1)}(\mathit{x})+\mathit{z}_{p_i}^{(2)}(\mathit{x})}{2}\right). 
\end{aligned}
\tag{9}\label{eq:vc}
\end{equation}

\noindent
Here, the distance between two distributions is normalized to be in the interval $[0,1]$, with a smaller value indicating higher similarity. Consequently, the objective $\mathcal{L}_{\mathit{vc}}$ of variational contrastive learning is to minimize this divergence for positive pairs, encouraging consistency between their latent representations.
\begin{equation}
	\small
	\begin{aligned}
		&\mathcal{L}_{\mathit{vc}}= \Big\{\min \left(\mathit{J S}\left(\mathit{z}_{p_i}^{(1)}(\mathit{x}), \mathit{z}_{p_i}^{(2)}(\mathit{x})\right)\right),\\
		& \left.\max \sum_{k=1, k \neq i}^{\mathit{B}} \mathit{J S}\left(\mathit{z}_{p_i}^{(1)}(\mathit{x}), \mathit{z}_{p_k}^{(1)}(\mathit{x})\right)
		+\mathit{J S}\left(\mathit{z}_{p_i}^{(1)}(\mathit{x}), \mathit{z}_{p_k}^{(2)}(\mathit{x})\right)\right.\Big\},
	\end{aligned}
	\tag{10}\label{eq:vc}
\end{equation}
\noindent
where $\mathit{B}$ denotes the number of samples in a batch.

In addition, from the definition of KL divergence, we have the following:

\vspace{-5pt}
\begin{equation}
	\small
	\begin{aligned}
		& \mathit{D}_{\mathit{KL}}\!\left(
		\mathit{z}_{p_i}^{(1)}(\mathit{x})
		\,\Big\|\,
		\frac{\mathit{z}_{p_i}^{(1)}(\mathit{x})+\mathit{z}_{p_i}^{(2)}(\mathit{x})}{2}
		\right)
		=\\
		& \int \mathit{z}_{p_i}^{(1)}(\mathit{x})
		\ln \mathit{z}_{p_i}^{(1)}(\mathit{x}) \, \mathrm{d}x
		-
		\int \mathit{z}_{p_i}^{(1)}(\mathit{x})
		\ln \frac{\mathit{z}_{p_i}^{(1)}(\mathit{x})+\mathit{z}_{p_i}^{(2)}(\mathit{x})}{2}
		\, \mathrm{d}x .
	\end{aligned}
	\tag{11}
\end{equation}
The JS divergence between two multivariate Gaussian distributions can now be formulated as follows:
\vspace{-5pt}
\begin{align}
	& \mathit{D}_{\mathit{JS}}\left( \mathit{z}_{p_i}^{(1)}(\mathit{x}),\mathit{z}_{p_i}^{(2)}(\mathit{x})\right) =-\frac{1}{4} \sum_{i=1}^{N} \text{In} \frac{\left(\sigma_{p_i}^{1} \sigma_{p_i}^{2}\right)^{2}}{\left(\frac{(\sigma_{p_i}^{1})^{2}+(\sigma_{p_i}^{2})^{2}}{2}\right)^{2}} \notag \\
	& +\frac{1}{4}\sum_{i=1}^{N}\frac{(\sigma_{p_i}^{1})^{2}+(\sigma_{p_i}^{2})^{2}+2\left(\frac{\mu_{p_i}^{1}-\mu_{p_i}^{2}}{2}\right)^{2}}{\left(\frac{(\sigma_{p_i}^{1})^{2}+(\sigma_{p_i}^{2})^{2}}{2}\right)^{2}} +2.
	\tag{12}
\end{align}
The JS divergence provides a convenient metric for quantifying the distance between two probability distributions. Inspired by the SimCLR framework \cite{DBLP:conf/cvpr/ChenH21}, we reformulate the contrastive distribution loss in Eq. \ref{eq:vc} as follows:
\begin{equation}
\begin{aligned}
	&\mathcal{L}_{\mathit{vc}}= \\
	&-\text{In}\frac{\mathit{e}^{\frac{-\mathit{D}_{\mathit{JS}}\left(\mathit{z}_{p_i}^{(1)}(\mathit{x}),\mathit{z}_{p_i}^{(2)}(\mathit{x})\right)}{
				\mathit{t}}}}{\sum_{k=1,k \neq i}^{N}\mathit{e}^{\frac{-\mathit{D}_{\mathit{JS}}\left(\mathit{z}_{p_i}^{(1)}(\mathit{x}),\mathit{z}_{p_k}^{(1)}(\mathit{x})\right)}{
				\mathit{t}}}+\mathit{e}^{\frac{-\mathit{D}_{\mathit{JS}}\left(\mathit{z}_{p_i}^{(1)}(\mathit{x}),\mathit{z}_{p_k}^{(2)}(\mathit{x})\right)}{
				\mathit{t}}}},
\end{aligned}
\tag{13}
\end{equation}
\noindent
where $N$ is the batch size and $\mathit{t}$ is the temperature parameter.

\paragraph{Generative Cross-Supervision Learning} As illustrated in Figure~\ref{fig:framework}, to encourage the encoder to capture invariant features, we incorporate a generative supervision signal into the framework. Specifically, latent embeddings $\mathit{g}_{p_i}^{(1)}$ and $\mathit{g}_{p_i}^{(2)}$ are sampled from the respective latent distributions $\mathit{z}_{p_i}^{(1)}(\mathit{x}) $ and $\mathit{z}_{p_i}^{(2)}(\mathit{x}) $. 
However, direct sampling from these distributions is non-differentiable, which hinders end-to-end training via backpropagation. To address this limitation, we adopt the reparameterization trick \cite{DBLP:journals/corr/KingmaW13} that enables differentiable sampling by expressing a stochastic variable as a deterministic function of the distribution parameters and a noise term:
\begin{equation}
\mathit{g}=\mu + \sigma \epsilon,
\tag{14}
\end{equation}
\noindent
where $\epsilon \sim \mathcal{N}(0,1)$, so that $\mathit{g} \sim \mathcal{N}(\mu, \sigma)$. The resulting samples $\mathit{g}_{p_i}^{(1)}$ and $\mathit{g}_{p_i}^{(2)}$ are then passed to a decoder with shared parameters. The decoder, implemented as a three-layer fully connected network with batch normalization, transforms the latent samples into reconstructed path embeddings $\hat{\mathit{z}}_{p_i}^{(1)}$ and $\hat{\mathit{z}}_{p_i}^{(2)}$. 
To guide this reconstruction, we employ a self-supervised objective. Conventional approaches typically use a direct supervision strategy, where the original embeddings $\mathit{z}_{p_i}^{(1)}$ and $\mathit{z}_{p_i}^{(2)} $ are used to supervise their respective reconstructions $\hat{\mathit{z}}_{p_i}^{(1)}$ and $\hat{\mathit{z}}_{p_i}^{(2)}$.
Instead, we introduce a cross-supervision mechanism, where  $\mathit{z}_{p_i}^{(1)}$ is used to supervise $\hat{\mathit{z}}_{p_i}^{(2)}$, and $\mathit{z}_{p_i}^{(2)}$ is used to supervise $\hat{\mathit{z}}_{p_i}^{(1)}$. This design offers two key advantages: (1) it forces the model to focus on semantically invariant features rather than spatially anchored or coordinate-dependent cues; and (2) as $\mathit{g}_{p_i}^{(1)}$ and $\mathit{g}_{p_i}^{(2)}$ are sampled from two distinct distributions, the cross-supervision strategy facilitates alignment between them, thereby reinforcing the core objective of variational contrastive learning.
The optimization objective of this module is formally defined as follows:
\begin{equation}
\mathcal{L}_{\mathit{cs}}=\max \left(\operatorname{sim}\left(\hat{\mathit{z}}_{p_i}^{(1)}, \mathit{z}_{p_i}^{(2)}\right)+\operatorname{sim}\left(\hat{\mathit{z}}_{p_i}^{(2)}, \mathit{z}_{p_i}^{(1)}\right)\right),
\tag{15}
\end{equation}
\noindent
where $\text{sim}(\cdot)$ denotes the similarity between two path embeddings and is computed via MSE.

As a result, the final objective is formulated as follows:

\begin{equation}
\mathcal{L}=\lambda_1 \mathcal{L}_\mathit{diff}+\lambda_2 \mathcal{L}_\mathit{vc}+\lambda_3\mathcal{L}_\mathit{cs},
\tag{16}
\end{equation}
\noindent
where $\lambda_1$, $\lambda_2$, and $\lambda_3$ are weight hyper-parameters that are used for balancing the contributions of the individual losses. Specifically, we adopt an uncertainty-based weighting strategy to dynamically adjust each loss component during training \cite{DBLP:conf/cvpr/KendallGC18}.

\subsection{Theoretical Analysis}

We next provide theoretical justification for the proposed generative
variational contrastive module. Let $C$ denote the identity of the underlying
path, $V\in\{1,2\}$ denote the augmentation-view index, and $G$ denote a
latent random variable sampled from the corresponding path distribution.

\paragraph{View Invariance Induced by Distributional Alignment.}

\noindent\textbf{Proposition 1.}
For a fixed path $C=i$, suppose that the two augmented views are sampled with
equal probability,
\begin{equation}
	\Pr(V=1)
	=
	\Pr(V=2)
	=
	\frac{1}{2}.
\end{equation}
Then,
\begin{equation}
	I(G;V\mid C=i)
	=
	D_{\mathrm{JS}}
	\left(
	P_i^{(1)},P_i^{(2)}
	\right).
	\label{eq:js_mutual_information}
\end{equation}

\noindent\textit{Proof.}
For a fixed path $C=i$, the marginal distribution of $G$ is
\begin{equation}
	M_i
	=
	\frac{1}{2}P_i^{(1)}
	+
	\frac{1}{2}P_i^{(2)}.
\end{equation}
Using the definition of conditional mutual information, we obtain
\begin{equation}
	\begin{aligned}
		I(G;V\mid C=i)
		&=
		\frac{1}{2}
		D_{\mathrm{KL}}\!\left(
		P_i^{(1)}\Vert M_i
		\right)
		\\
		&\quad+
		\frac{1}{2}
		D_{\mathrm{KL}}\!\left(
		P_i^{(2)}\Vert M_i
		\right)
		\\
		&=
		D_{\mathrm{JS}}\!\left(
		P_i^{(1)},P_i^{(2)}
		\right).
	\end{aligned}
	\label{eq:js_mutual_information_proof}
\end{equation}
Therefore,
\begin{equation}
	D_{\mathrm{JS}}
	\left(
	P_i^{(1)},P_i^{(2)}
	\right)
	=
	0
\end{equation}
implies
\begin{equation}
	I(G;V\mid C=i)
	=
	0,
\end{equation}
and hence
\begin{equation}
	G
	\perp
	V
	\mid
	C=i.
\end{equation}
Thus, minimizing the positive-pair JS divergence directly suppresses
augmentation-specific information in the latent representation.
\hfill$\square$

This proposition gives an information-theoretic interpretation of
distributional alignment. Rather than merely forcing two augmented
representations to be numerically close, the proposed objective explicitly
reduces the information that the latent representation contains about which
augmentation generated it.

Moreover, the JS divergence also controls the discrepancy between the two
view-dependent distributions. Let
$\operatorname{TV}(P,Q)$ denote the total variation distance. Applying
Pinsker's inequality to the two KL terms in the JS divergence yields
\begin{equation}
	\operatorname{TV}
	\left(
	P_i^{(1)},P_i^{(2)}
	\right)
	\leq
	\sqrt{
		2
		D_{\mathrm{JS}}
		\left(
		P_i^{(1)},P_i^{(2)}
		\right)
	}.
	\label{eq:tv_js_bound}
\end{equation}
Therefore, for any measurable function $h$ satisfying
$0\leq h(G)\leq 1$,
\begin{equation}
	\begin{aligned}
		&
		\left|
		\mathbb{E}_{P_i^{(1)}}[h(G)]
		-
		\mathbb{E}_{P_i^{(2)}}[h(G)]
		\right|
		\\
		&\qquad
		\leq
		\sqrt{
			2
			D_{\mathrm{JS}}
			\left(
			P_i^{(1)},P_i^{(2)}
			\right)
		}.
	\end{aligned}
	\label{eq:prediction_bound}
\end{equation}
Hence, decreasing the JS divergence also reduces an upper bound on the
prediction discrepancy between different views of the same path.

\paragraph{Distributional Margin and Path Discriminability.}

\noindent\textbf{Proposition 2.}
Consider an anchor distribution $P_i^{(1)}$, its positive distribution
$P_i^{(2)}$, and $M$ negative distributions
$\{P_j^{-}\}_{j=1}^{M}$. Define
\begin{equation}
	d_i^{+}
	=
	D_{\mathrm{JS}}
	\left(
	P_i^{(1)},P_i^{(2)}
	\right)
\end{equation}
and
\begin{equation}
	d_{ij}^{-}
	=
	D_{\mathrm{JS}}
	\left(
	P_i^{(1)},P_j^{-}
	\right).
\end{equation}
The corresponding contrastive objective is
\begin{equation}
	\ell_i
	=
	-\log
	\frac{
		\exp(-d_i^{+}/t)
	}{
		\exp(-d_i^{+}/t)
		+
		\sum_{j=1}^{M}
		\exp(-d_{ij}^{-}/t)
	}.
	\label{eq:theoretical_contrastive}
\end{equation}
Suppose that
\begin{equation}
	d_i^{+}
	\leq
	\epsilon
\end{equation}
and
\begin{equation}
	d_{ij}^{-}
	\geq
	\delta,
	\qquad
	\forall j,
\end{equation}
where $\delta>\epsilon$. Then,
\begin{equation}
	\ell_i
	\leq
	\log
	\left[
	1
	+
	M
	\exp
	\left(
	-\frac{\delta-\epsilon}{t}
	\right)
	\right].
	\label{eq:distributional_margin_bound}
\end{equation}

\noindent\textit{Proof.}
Equation~\eqref{eq:theoretical_contrastive} can be rewritten as
\begin{equation}
	\ell_i
	=
	\log
	\left[
	1
	+
	\sum_{j=1}^{M}
	\exp
	\left(
	-\frac{
		d_{ij}^{-}
		-
		d_i^{+}
	}{t}
	\right)
	\right].
\end{equation}
Since
\begin{equation}
	d_{ij}^{-}
	-
	d_i^{+}
	\geq
	\delta-\epsilon,
\end{equation}
we obtain
\begin{equation}
	\ell_i
	\leq
	\log
	\left[
	1
	+
	M
	\exp
	\left(
	-\frac{\delta-\epsilon}{t}
	\right)
	\right].
\end{equation}
\hfill$\square$

Proposition~2 indicates that minimizing the contrastive objective implicitly
maximizes the distributional margin
\begin{equation}
	\Delta
	=
	\delta-\epsilon
\end{equation}
between positive and negative pairs. As $\Delta$ increases, the upper bound
in Eq.~\eqref{eq:distributional_margin_bound} decreases exponentially.
Therefore, the proposed objective simultaneously promotes intra-path
compactness and inter-path separation.

The contrastive objective also admits an information-theoretic
interpretation. Let
\begin{equation}
	R^{(v)}
	=
	\left(
	\boldsymbol{\mu}^{(v)},
	\boldsymbol{\sigma}^{(v)}
	\right)
\end{equation}
denote the distributional representation of the $v$-th view. Under the
standard negative-sampling setting of InfoNCE, the mutual information between
the two views satisfies
\begin{equation}
	I
	\left(
	R^{(1)};R^{(2)}
	\right)
	\geq
	\log K
	-
	\mathbb{E}
	\left[
	\mathcal{L}_{\mathrm{vc}}
	\right],
	\label{eq:infonce_bound}
\end{equation}
where $K$ denotes the number of candidate samples in the contrastive set.

Furthermore, suppose that the two augmented representations are conditionally
independent given their underlying path identity $C$, i.e.,
\begin{equation}
	R^{(1)}
	-
	C
	-
	R^{(2)}
\end{equation}
forms a Markov chain. By the data-processing inequality,
\begin{equation}
	I
	\left(
	R^{(1)};R^{(2)}
	\right)
	\leq
	I
	\left(
	C;R^{(1)}
	\right)
\end{equation}
and similarly
\begin{equation}
	I
	\left(
	R^{(1)};R^{(2)}
	\right)
	\leq
	I
	\left(
	C;R^{(2)}
	\right).
\end{equation}
Consequently, minimizing $\mathcal{L}_{\mathrm{vc}}$ increases a lower bound
on the information shared between the two augmented representations, thereby
encouraging both views to retain information associated with the underlying
path identity.

\paragraph{Invariant Feature Learning via Cross-Supervision.}

\noindent\textbf{Proposition 3.}
Assume that the representations of two augmented path views can be decomposed
as
\begin{equation}
	Z^{(1)}
	=
	S
	+
	N^{(1)},
	\qquad
	Z^{(2)}
	=
	S
	+
	N^{(2)},
	\label{eq:semantic_nuisance_decomposition}
\end{equation}
where $S$ denotes the semantic information shared by the two views, and
$N^{(1)}$ and $N^{(2)}$ denote view-specific nuisance components.
Assume that
\begin{equation}
	\mathbb{E}
	\left[
	N^{(v)}
	\mid
	S
	\right]
	=
	0,
	\qquad
	v\in\{1,2\},
	\label{eq:zero_mean_nuisance}
\end{equation}
and that the nuisance components of different views are conditionally
independent given $S$.

Let $G^{(1)}$ denote a latent representation generated from the first view.
Consider the cross-reconstruction risk
\begin{equation}
	\mathcal{R}_{\mathrm{cross}}(D)
	=
	\mathbb{E}
	\left[
	\left\|
	D(G^{(1)})
	-
	Z^{(2)}
	\right\|_2^2
	\right].
	\label{eq:cross_reconstruction_risk}
\end{equation}
Then, the optimal decoder under the squared-error criterion satisfies
\begin{equation}
	D^{*}(G^{(1)})
	=
	\mathbb{E}
	\left[
	S
	\mid
	G^{(1)}
	\right].
	\label{eq:optimal_cross_decoder}
\end{equation}

\noindent\textit{Proof.}
Under the mean squared error criterion, the optimal predictor is the
conditional expectation:
\begin{equation}
	D^{*}(G^{(1)})
	=
	\mathbb{E}
	\left[
	Z^{(2)}
	\mid
	G^{(1)}
	\right].
\end{equation}
Substituting Eq.~\eqref{eq:semantic_nuisance_decomposition}, we obtain
\begin{align}
	D^{*}(G^{(1)})
	&=
	\mathbb{E}
	\left[
	S
	+
	N^{(2)}
	\mid
	G^{(1)}
	\right]
	\nonumber\\
	&=
	\mathbb{E}
	\left[
	S
	\mid
	G^{(1)}
	\right]
	+
	\mathbb{E}
	\left[
	N^{(2)}
	\mid
	G^{(1)}
	\right].
\end{align}
Under the conditional-independence and zero-mean assumptions,
\begin{equation}
	\mathbb{E}
	\left[
	N^{(2)}
	\mid
	G^{(1)}
	\right]
	=
	0.
\end{equation}
Therefore,
\begin{equation}
	D^{*}(G^{(1)})
	=
	\mathbb{E}
	\left[
	S
	\mid
	G^{(1)}
	\right].
\end{equation}
\hfill$\square$

The minimum cross-reconstruction risk can further be decomposed as
\begin{align}
	\min_D
	\mathcal{R}_{\mathrm{cross}}(D)
	&=
	\mathbb{E}
	\left[
	\left\|
	S
	-
	\mathbb{E}
	\left[
	S
	\mid
	G^{(1)}
	\right]
	\right\|_2^2
	\right]
	\nonumber\\
	&\quad
	+
	\mathbb{E}
	\left[
	\left\|
	N^{(2)}
	\right\|_2^2
	\right].
	\label{eq:cross_risk_decomposition}
\end{align}
The second term is independent of the encoder and therefore represents an
irreducible component. Hence, reducing the trainable part of the
cross-reconstruction loss requires improving the representation of the shared
semantic component $S$.

In contrast, conventional self-reconstruction minimizes
\begin{equation}
	\mathcal{R}_{\mathrm{self}}(D)
	=
	\mathbb{E}
	\left[
	\left\|
	D(G^{(1)})
	-
	Z^{(1)}
	\right\|_2^2
	\right].
\end{equation}
Because $Z^{(1)}$ contains both $S$ and $N^{(1)}$, the encoder may reduce the
self-reconstruction error by preserving view-specific nuisance information.
Cross-supervision removes this shortcut because the nuisance component of one
view is not predictive of the nuisance component of another view.

\paragraph{Overall Theoretical Interpretation.}
The above results reveal the complementary roles of the proposed objectives.
Proposition~1 shows that distributional alignment directly minimizes
\begin{equation}
	I(G;V\mid C),
\end{equation}
thereby suppressing augmentation-specific information. Proposition~2 shows
that distributional contrast promotes a large inter-path distributional
margin while preserving information shared across views of the same path.
Proposition~3 demonstrates that generative cross-supervision preferentially
preserves semantic information shared between augmented views while
suppressing view-specific nuisance factors.

Therefore, jointly optimizing
\begin{equation}
	\mathcal{L}
	=
	\lambda_{1}\mathcal{L}_{\mathrm{diff}}
	+
	\lambda_{2}\mathcal{L}_{\mathrm{vc}}
	+
	\lambda_{3}\mathcal{L}_{\mathrm{cs}}
\end{equation}
encourages the learned path representations to be simultaneously
view-invariant, path-discriminative, and semantically informative.

\section{Experiments}
\noindent
\paragraph{Datasets}We conduct experiments on three real-world datasets: Aalborg (Denmark) \cite{DBLP:journals/tkde/YangGY22}, Chengdu (China) \cite{DBLP:journals/pvldb/00020B22}, and Harbin (China) \cite{DBLP:conf/www/LiCSC19}. Each dataset includes a road network extracted from OpenStreetMap and GPS trajectories. The Aalborg road network consists of 10,017 nodes and 11,597 edges; Chengdu comprises 6,632 nodes and 17,038 edges; and Harbin contains 8,497 nodes and 14,497 edges. After map-matching \cite{DBLP:journals/gis/YangG18,DBLP:conf/aaai/HanYH26,DBLP:journals/corr/abs-2601-08482}, we obtain 33,264 paths for Aalborg, 54,491 for Chengdu, and 259,615 for Harbin.

\begin{table*}[t]
\small

\centering
{\renewcommand{\arraystretch}{1}
\begin{tabular}{ll|lll|lll|lll}
	\toprule[1.5pt]
	\multicolumn{2}{c|}{\multirow{2}{*}{\textbf{Method}}}                             & \multicolumn{3}{c|}{\textbf{Aalborg}}                                                  & \multicolumn{3}{c|}{\textbf{Chengdu}}                                                  & \multicolumn{3}{c}{\textbf{Harbin}}                                                   \\ \cline{3-11} 
	\multicolumn{2}{l|}{}                                                    & \multicolumn{1}{l}{\textbf{MAE}$\downarrow$} & \multicolumn{1}{l}{\textbf{MAPE}(\%)$\downarrow$} & \multicolumn{1}{l|}{\textbf{RMSE}$\downarrow$} & \multicolumn{1}{l}{\textbf{MAE}$\downarrow$} & \multicolumn{1}{l}{\textbf{MAPE}(\%)$\downarrow$} & \multicolumn{1}{l|}{\textbf{RMSE}$\downarrow$} & \multicolumn{1}{l}{\textbf{MAE}$\downarrow$} & \multicolumn{1}{l}{\textbf{MAPE}(\%)$\downarrow$} &  \multicolumn{1}{l}{\textbf{RMSE}$\downarrow$ }\\ \toprule[0.9pt]
	\multicolumn{1}{l|}{\multirow{4}{*}{\rotatebox[origin=c]{90}{Supervised}}}   
	& \textsc{PathRank}      & \multicolumn{1}{c}{2.913}             & \multicolumn{1}{c}{18.713}              &  \multicolumn{1}{c|}{4.832}              & \multicolumn{1}{c}{5.802}             & \multicolumn{1}{c}{34.014}              & \multicolumn{1}{c|}{8.894}               & \multicolumn{1}{c}{4.127}             & \multicolumn{1}{c}{14.339}              & \multicolumn{1}{c}{5.301}               \\ \cline{2-11} 
	
	\multicolumn{1}{l|}{}                              & \textsc{HierETA} & \multicolumn{1}{c}{3.097}             & \multicolumn{1}{c}{18.620}              & \multicolumn{1}{c|}{5.058}            & \multicolumn{1}{c}{5.060}             & \multicolumn{1}{c}{28.820}              & \multicolumn{1}{c|}{7.921}               & \multicolumn{1}{c}{4.897}             & \multicolumn{1}{c}{16.506}              &\multicolumn{1}{c}{5.961}                \\ \cline{2-11} 

	\multicolumn{1}{l|}{}                              & \textsc{CompactETA} & \multicolumn{1}{c}{3.030}             & \multicolumn{1}{c}{19.726}              & \multicolumn{1}{c|}{4.791}              & \multicolumn{1}{c}{\underline{4.832}}             & \multicolumn{1}{c}{28.678}              & \multicolumn{1}{c|}{7.447}              & \multicolumn{1}{c}{4.372}             & \multicolumn{1}{c}{14.974}              &  \multicolumn{1}{c}{5.362}              \\ \cline{2-11} 
	
	\multicolumn{1}{l|}{}                              & \textsc{HMTRL}    & \multicolumn{1}{c}{3.515}             & \multicolumn{1}{c}{19.955}              &\multicolumn{1}{c|}{5.792 }              & \multicolumn{1}{c}{4.861}             & \multicolumn{1}{c}{28.529}              &\multicolumn{1}{c|}{7.473}               & \multicolumn{1}{c}{4.704}             & \multicolumn{1}{c}{15.543}              & \multicolumn{1}{c}{5.803}                     \\ \toprule[0.9pt]
	
	\multicolumn{1}{l|}{\multirow{8}{*}{\rotatebox[origin=c]{90}{Unsupervised}}} & \textsc{Node2vec}   & \multicolumn{1}{c}{5.701}       & \multicolumn{1}{c}{35.485}          &  \multicolumn{1}{c|}{8.165}         & \multicolumn{1}{c}{5.997}       & \multicolumn{1}{c}{35.919}          &  \multicolumn{1}{c|}{8.653}         & \multicolumn{1}{c}{5.399}       & \multicolumn{1}{c}{18.709}          & \multicolumn{1}{c}{6.773}         \\ \cline{2-11} 
	
	\multicolumn{1}{l|}{}                              & \textsc{Toast}      & \multicolumn{1}{c}{3.439}             & \multicolumn{1}{c}{21.311}              &\multicolumn{1}{c|}{5.555}               & \multicolumn{1}{c}{4.918}             & \multicolumn{1}{c}{28.567}              &\multicolumn{1}{c|}{7.513}               & \multicolumn{1}{c}{4.627}             & \multicolumn{1}{c}{16.101}              &\multicolumn{1}{c}{5.803}               \\ \cline{2-11} 

	\multicolumn{1}{l|}{}                              & \textsc{PIM}        & \multicolumn{1}{c}{3.471}             & \multicolumn{1}{c}{21.314}              &\multicolumn{1}{c|}{5.640}               & \multicolumn{1}{c}{4.924}             & \multicolumn{1}{c}{28.553}              &\multicolumn{1}{c|}{7.493}               & \multicolumn{1}{c}{4.538}             & \multicolumn{1}{c}{15.713}              &\multicolumn{1}{c}{5.682}               \\ \cline{2-11} 
	
	\multicolumn{1}{l|}{}                              & \textsc{START}      & \multicolumn{1}{c}{2.965}             & \multicolumn{1}{c}{18.432}              &\multicolumn{1}{c|}{4.979}               & \multicolumn{1}{c}{5.174}             & \multicolumn{1}{c}{30.769}              &\multicolumn{1}{c|}{7.726}                 & \multicolumn{1}{c}{4.402}             & \multicolumn{1}{c}{15.262}              &\multicolumn{1}{c}{5.503}                 \\ \cline{2-11} 
	
	\multicolumn{1}{l|}{}                              & \textsc{LightPath}  & \multicolumn{1}{c}{2.920}             & \multicolumn{1}{c}{18.254}              &\multicolumn{1}{c|}{4.932}               & \multicolumn{1}{c}{4.850}             & \multicolumn{1}{c}{28.387}              &\multicolumn{1}{c|}{7.406}               & \multicolumn{1}{c}{4.221}             & \multicolumn{1}{c}{14.962}              &\multicolumn{1}{c}{5.310}               \\ \cline{2-11} 
	
	\multicolumn{1}{l|}{}                              & \textsc{RED}       & \multicolumn{1}{c}{\underline{2.457}}             & \multicolumn{1}{c}{\underline{14.907}}              &\multicolumn{1}{c|}{\underline{3.809}}               & \multicolumn{1}{c}{\underline{4.845}}             & \multicolumn{1}{c}{\underline{28.380}}              &\multicolumn{1}{c|}{\underline{7.296}}                 & \multicolumn{1}{c}{\underline{4.097}}             & \multicolumn{1}{c}{\underline{14.214}}              &\multicolumn{1}{c}{\underline{5.153}}                 \\ \cline{2-11} 
	
	\multicolumn{1}{l|}{}                              & \textsc{DGCPath}      & \multicolumn{1}{c}{\textbf{2.131}}             & \multicolumn{1}{c}{\textbf{13.654}}              &\multicolumn{1}{c|}{\textbf{3.725}}               & \multicolumn{1}{c}{\textbf{4.659}}             & \multicolumn{1}{c}{\textbf{27.488}}              &\multicolumn{1}{c|}{\textbf{7.064}}                 & \multicolumn{1}{c}{\textbf{4.016}}             & \multicolumn{1}{c}{\textbf{13.905}}              &\multicolumn{1}{c}{\textbf{5.042}}                 \\ \bottomrule[1.5pt]
	
\end{tabular}}
\caption{{Overall Accuracy at Travel Time Estimation. }}
\label{tab:miantte}
\vspace{-5pt} 
\end{table*}

\begin{table*}[htp]

\centering
\small
{\renewcommand{\arraystretch}{1}
\begin{tabular}{ll|lll|lll|lll}
	\toprule[1.5pt]
	\multicolumn{2}{c|}{\multirow{2}{*}{\textbf{Method}}}                             & \multicolumn{3}{c|}{\textbf{Aalborg}}                                                  & \multicolumn{3}{c|}{\textbf{Chengdu}}                                                  & \multicolumn{3}{c}{\textbf{Harbin}}                                                   \\ \cline{3-11} 
	\multicolumn{2}{l|}{}                                                    & \multicolumn{1}{l}{\textbf{MAE}$\downarrow$} & \multicolumn{1}{l}{\textbf{TAU}($\tau$)$\uparrow$} & \textbf{RHO}($\rho$)$\uparrow$ & \multicolumn{1}{l}{\textbf{MAE}$\downarrow$} & \multicolumn{1}{l}{\textbf{TAU}($\tau$)$\uparrow$} & \textbf{RHO}($\rho$)$\uparrow$ & \multicolumn{1}{l}{\textbf{MAE}$\downarrow$} & \multicolumn{1}{l}{\textbf{TAU}($\tau$)$\uparrow$} & \textbf{RHO}($\rho$)$\uparrow$ \\ \toprule[0.9pt]
	\multicolumn{1}{l|}{\multirow{4}{*}{\rotatebox[origin=c]{90}{Supervised}}}    & \textsc{PathRank}   & \multicolumn{1}{c}{0.159}             & \multicolumn{1}{c}{0.727}              &\multicolumn{1}{c|}{0.751}               & \multicolumn{1}{c}{\underline{0.123}}             & \multicolumn{1}{c}{0.703}              &\multicolumn{1}{c|}{0.742}                 & \multicolumn{1}{c}{0.145}             & \multicolumn{1}{c}{0.821}              &\multicolumn{1}{c}{0.832}                 \\ \cline{2-11} 
	
	\multicolumn{1}{l|}{}                               & \textsc{HierETA}   & \multicolumn{1}{c}{0.152}             & \multicolumn{1}{c}{0.757}              &\multicolumn{1}{c|}{0.767}               & \multicolumn{1}{c}{0.178}             & \multicolumn{1}{c}{0.735}              &\multicolumn{1}{c|}{0.774}                 & \multicolumn{1}{c}{0.133}             & \multicolumn{1}{c}{0.846}              &\multicolumn{1}{c}{0.879}               \\ \cline{2-11} 
	
	\multicolumn{1}{l|}{}                               & \textsc{CompactETA}   & \multicolumn{1}{c}{0.169}             & \multicolumn{1}{c}{0.698}              &\multicolumn{1}{c|}{0.709}               & \multicolumn{1}{c}{0.195}             & \multicolumn{1}{c}{0.661}              &\multicolumn{1}{c|}{0.705}                 & \multicolumn{1}{c}{0.160}             & \multicolumn{1}{c}{0.796}              &\multicolumn{1}{c}{0.833}               \\ \cline{2-11} 
	
	\multicolumn{1}{l|}{}                               & \textsc{HMTRL}   & \multicolumn{1}{c}{0.163}             & \multicolumn{1}{c}{0.729}              &\multicolumn{1}{c|}{0.741}               & \multicolumn{1}{c}{0.149}             & \multicolumn{1}{c}{0.779}              &\multicolumn{1}{c|}{0.789}                 & \multicolumn{1}{c}{0.133}             & \multicolumn{1}{c}{0.838}              &\multicolumn{1}{c}{0.874}                        \\ \toprule[0.9pt]

	\multicolumn{1}{l|}{\multirow{8}{*}{\rotatebox[origin=c]{90}{Unsupervised}}}  & \textsc{Node2vec}      & \multicolumn{1}{c}{0.253}             & \multicolumn{1}{c}{0.632}              &\multicolumn{1}{c|}{0.637}               & \multicolumn{1}{c}{0.246}             & \multicolumn{1}{c}{0.464}              &\multicolumn{1}{c|}{0.505}                 & \multicolumn{1}{c}{0.211}             & \multicolumn{1}{c}{0.468}              &\multicolumn{1}{c}{0.512}                  \\ \cline{2-11}  
	
	\multicolumn{1}{l|}{}                              & \textsc{Toast}      & \multicolumn{1}{c}{0.138}             & \multicolumn{1}{c}{0.768}              &\multicolumn{1}{c|}{0.779}               & \multicolumn{1}{c}{0.131}             & \multicolumn{1}{c}{0.734}              &\multicolumn{1}{c|}{0.752}                 & \multicolumn{1}{c}{0.161}             & \multicolumn{1}{c}{0.840}              &\multicolumn{1}{c}{0.878}                \\ \cline{2-11} 
	
	\multicolumn{1}{l|}{}                              & \textsc{PIM}      & \multicolumn{1}{c}{0.141}             & \multicolumn{1}{c}{0.764}              &\multicolumn{1}{c|}{0.777}               & \multicolumn{1}{c}{0.125}             & \multicolumn{1}{c}{0.725}              &\multicolumn{1}{c|}{0.745}                 & \multicolumn{1}{c}{\underline{0.119}}             & \multicolumn{1}{c}{0.879}              &\multicolumn{1}{c}{0.911}                 \\ \cline{2-11} 
	
	\multicolumn{1}{l|}{}                              & \textsc{START}      & \multicolumn{1}{c}{0.178}             & \multicolumn{1}{c}{\underline{0.839}}              &\multicolumn{1}{c|}{\underline{0.859}}               & \multicolumn{1}{c}{0.228}             & \multicolumn{1}{c}{0.638}              &\multicolumn{1}{c|}{0.679}                 & \multicolumn{1}{c}{0.151}             & \multicolumn{1}{c}{0.859}              &\multicolumn{1}{c}{0.890}                 \\ \cline{2-11} 
	
	\multicolumn{1}{l|}{}                              & \textsc{LightPath}  & \multicolumn{1}{c}{0.118}             & \multicolumn{1}{c}{0.781}              &\multicolumn{1}{c|}{0.791}               & \multicolumn{1}{c}{0.195}             & \multicolumn{1}{c}{0.711}              &\multicolumn{1}{c|}{0.748}                 & \multicolumn{1}{c}{0.122}             & \multicolumn{1}{c}{\underline{0.881}}              &\multicolumn{1}{c}{\underline{0.913}}                 \\ \cline{2-11} 
	
	\multicolumn{1}{l|}{}                              & \textsc{RED}        & \multicolumn{1}{c}{\underline{0.109}}             & \multicolumn{1}{c}{0.819}              &\multicolumn{1}{c|}{0.823}               & \multicolumn{1}{c}{0.129}             & \multicolumn{1}{c}{\underline{0.797}}              &\multicolumn{1}{c|}{\underline{0.833}}                 & \multicolumn{1}{c}{0.127}             & \multicolumn{1}{c}{0.858}              &\multicolumn{1}{c}{0.865}                 \\ \cline{2-11} 

	\multicolumn{1}{l|}{}                             & \textsc{DGCPath}            & \multicolumn{1}{c}{\textbf{0.107}}             & \multicolumn{1}{c}{\textbf{0.846}}              &\multicolumn{1}{c|}{\textbf{0.870}}               & \multicolumn{1}{c}{\textbf{0.114}}             & \multicolumn{1}{c}{\textbf{0.819}}              &\multicolumn{1}{c|}{\textbf{0.859} }                & \multicolumn{1}{c}{\textbf{0.112}}             & \multicolumn{1}{c}{\textbf{0.898}}              &\multicolumn{1}{c}{\textbf{0.925}}                  \\ \bottomrule[1.5pt]

\end{tabular}}
\caption{{Overall Accuracy at Path Ranking. }}
\label{tab:mianpr}
\vspace{-5pt}
\end{table*}

\paragraph{Downstream Tasks}\textit{Path Travel Time Estimation: }
Each path is associated with a travel time (seconds) obtained from trajectories. 
We aim to build a regression model to estimate the travel times of paths. 
We evaluate the accuracy of the estimations using 
Mean Absolute Error (MAE), Mean Absolute Percentage Error (MAPE), and Root Mean Square Error (RMSE). Smaller values indicate higher estimation accuracy.  

\noindent
\textit{Path Ranking: }
Given a set of candidate paths---typically sharing the same source and destination---each path is assigned a ranking score within the range $[0,1]$, reflecting its relative preference based on historical trajectory data. Follow previous studies \cite{DBLP:journals/tkde/YangGY22,DBLP:conf/icde/Yang020,DBLP:journals/corr/abs-1907-04028}, we compute these scores using similarity measurements derived from real-world trajectories. To evaluate the performance of path ranking, we use three widely used metrics: Mean Absolute Error (MAE), the Kendall rank correlation coefficient ($\tau$), and the Sperman rank correlation coefficient ($\rho$). These metrics collectively assess both the accuracy of the predicted ranking scores and the consistency of the predicted orderings relative to the ground truth.

\paragraph{Baselines} We consider 10 baseline methods, including 6 unsupervised and 4 supervised methods. The unsupervised methods include
\textsc{Node2vec}~\cite{DBLP:conf/kdd/GroverL16},  \textsc{Toast}~\cite{DBLP:conf/cikm/ChenLCBLLCE21},
\textsc{PIM}~\cite{DBLP:conf/ijcai/YangGHT021},
\textsc{START}~\cite{DBLP:conf/icde/JiangPRJLW23},
\textsc{LightPath}~\cite{DBLP:conf/kdd/YangHGYJ23}, and 
\textsc{RED}~\cite{DBLP:journals/pvldb/ZhouSCJK24}. Among them, \textsc{PIM}, \textsc{START}, and \textsc{LightPath} rely on pre-defined view augmentation strategies, such as random masking or node dropping.
The supervised methods include {PathRank}~\cite{DBLP:journals/tkde/YangGY22},
\textsc{HierETA}~\cite{DBLP:conf/kdd/ChenXGFMCC22}, 
\textsc{CompactETA}~\cite{DBLP:conf/kdd/FuMYW20},  and
\textsc{HMTRL}~\cite{DBLP:journals/vldb/LiuHFLCX23}. These methods leverage task-specific labels to learn path representations.

For all unsupervised methods, we first pre-train the path encoder on unlabeled trajectory data (i.e., 30/50/200K samples from Aalborg/Chengdu/Harbin)
to obtain path embeddings. These embeddings are subsequently fed to a regression model trained on a smaller labeled subset (e.g., 12K labeled paths from Aalborg) to predict travel time and path ranking scores. We adopt the Gradient Boosting Regressor (GBR)~\cite{DBLP:conf/nips/PeterDHN17} as the predictive model due to the regression nature of our tasks. In contrast, supervised methods train the path encoder end-to-end using only the labeled data (e.g., 12K samples per city) without leveraging unlabeled paths.


\paragraph{Impementation Details}
We employ a Transformer architecture and
randomly initialize all learnable parameters with uniform distributions. We employ node2vec to embed each node to 128-dimensional vectors and set the dimension for path representation to 128. For a fair comparison, we set the
path representation dimensionality of all baseline methods as 128. We use the AdamW optimizer with a cosine decay learning rate
schedule over 600 epochs, with a warm-up period of 50 epochs. We set the base learning rate to 1e-5 for Aalborg and Chengdu datasets, and 1e-6 for the Harbin dataset. The source code is available at \url{https://github.com/Sean-Bin-Yang/DGCPath}.
\subsection{Performance Findings}
We first evaluate the accuracy of all methods on the downstream tasks.
Tables \ref{tab:miantte} and \ref{tab:mianpr} show the performance of all methods at travel time estimation and path ranking tasks on the three datasets. The overall best performance is marked in bold, and the second-best performance is underlined. We make three main observations:
1) \textsc{DGCPath} achieves the best performance across all baselines at both tasks on all three datasets. Specifically, at the travel time estimation task on the Aalborg dataset, \textsc{DGCPath} reduces MAE and RMSE by up to 13.27\% and 2.21\%, respectively. At the path ranking task on the Harbin dataset, it increases Kendall’s $\tau$ and Spearman’s $\rho$ by up to 4.66\% and 6.94\%, respectively. These improvements stem from the improved feature learning capabilities of the proposed diffusion-based view generator. Furthermore, the use of a distribution-aware generative contrastive loss enables the model to learn more generalizable path representations, leading to enhanced overall performance. 
2) The supervised methods, including \textsc{PathRank}, \textsc{HierETA}, \textsc{CompactETA}, and \textsc{HMTRL}, show relatively poor performance, primarily due to the limited size of the labeled training data. Collecting task-specific labels is often expensive and labor-intensive. The experiments are thus performed in a realistic setting where labeled data is scarce. 
3) All unsupervised methods are capable of learning generic path representations. Methods like \textsc{Toast}, \textsc{PIM}, \textsc{START}, \textsc{LightPath}, and \textsc{RED} consistently outperform \textsc{Node2vec} across all three datasets. This is because \textsc{Node2vec}, originally designed for node-level representation, fails to capture the sequential dependencies between nodes in a path. In contrast, the other unsupervised methods are designed specifically for path representation learning and recognize the importance of effectively modeling spatial dependencies. Among them, \textsc{DGCPath} achieves superior performance due to its integration of a diffusion-based view generator and a distribution-aware contrastive learning framework, enabling it to better capture both spatial structure and representation diversity.

\begin{figure}[tp]
	\centering
	\includegraphics[scale=0.42]{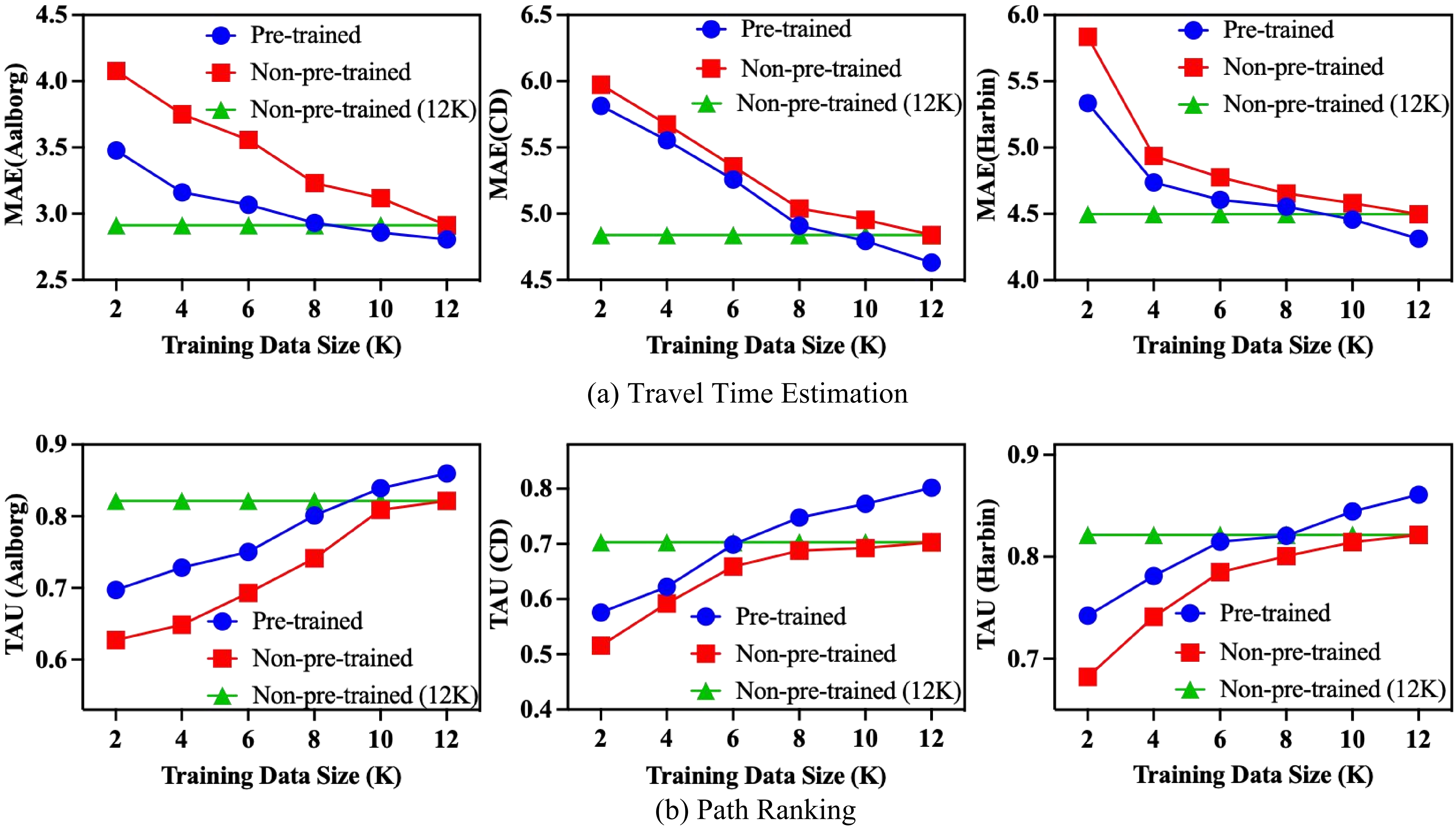}
	\caption{Effects of Pre-training. }
	\label{fig:EP}
	\vspace{-10pt}
\end{figure}

\subsection{Ablation Study}
We study three additional variations of our method to evaluate the contribution of each key component: (1) \emph{w/o DVG} excludes the diffusion-based view generator, (2) \emph{w/o VCM} excludes the variational contrastive module, and (3) \emph{w/o GCSM} excludes the generative cross-supervision module.
The results, in Table~\ref{tb:joint}, show that all components contribute to the performance of \textsc{DGCPath} at both tasks, as removing any of the components yields a performance drop. However, the extent of each component’s impact varies by task. Notably, the \emph{VCM} plays a more critical role than \emph{DVG} and \emph{GCSM}, indicating its importance in learning generalizable path representations. Removing \emph{DVG} or \emph{GCSM} also degrades performance, highlighting the complementary strengths of all modules.

\subsection{Effects of Pre-training}

To further assess the benefits of the proposed method, we use \textsc{DGCPath} to pre-train the Transformer-based path encoder for the supervised model {PathRank}. Specifically, \textsc{DGCPath} is first trained in an unsupervised manner, after which the learned parameters are transferred to initialize the Transformer encoder in \textsc{PathRank}.
Figure~\ref{fig:EP} presents the performance of \textsc{PathRank} with and without pre-training at travel time estimation and path ranking, under varying labeled data sizes: 2K, 4K, 6K, 8K, 10K, and 12K. We observe that
with pre-training, \textsc{PathRank} achieves comparable performance to its non-pretrained counterpart using significantly fewer labeled training samples. For instance, at the travel time estimation task, \textsc{PathRank} with pre-training requires only 8K, 9K, and 9K labeled samples on Aalborg, Chengdu, and Harbin, respectively, to match the performance of \textsc{PathRank} trained from scratch with 12K labeled samples. When \textsc{PathRank} is trained using the full 12K labeled samples with pre-training, it outperforms the non-pretrained counterpart.
\begin{table}[tp]

\centering
\begin{tabular}{l|ll}
	\toprule[1.5pt]
	\multirow{3}{*}{\textbf{Methods}} & \multicolumn{2}{c}{\textbf{Aalborg}}                                          \\ \cline{2-3} 
	& \multicolumn{1}{c|}{\textbf{Travel Time Estimation}} & \textbf{Path Ranking}   \\ \cline{2-3} 
	& \multicolumn{1}{l|}{\textbf{MAE/MAPE/RMSE}}          & \textbf{MAE/$\tau$/$\rho$}    \\ \toprule[0.9pt]
	\emph{w/o DVG}                           & \multicolumn{1}{c|}{2.33/14.23/3.89}                 & 0.12/0.80/0.81          \\ \hline
	\emph{w/o VCM}                           & \multicolumn{1}{c|}{2.64/16.05/4.75}                 & 0.13/0.77/0.80          \\ \hline
	\emph{w/o GCSM}                          & \multicolumn{1}{c|}{2.31/14.74/4.01}                 & 0.14/0.83/0.84          \\ \hline
	\textsc{DGCPath}                           & \multicolumn{1}{c|}{\textbf{2.13/13.65/3.73}}                 & \textbf{0.11/0.85/0.87}          \\ \toprule[1.5pt]
	\multirow{3}{*}{\textbf{Methods}} & \multicolumn{2}{c}{\textbf{Chengdu}}                                          \\ \cline{2-3} 
	& \multicolumn{1}{c|}{\textbf{Travel Time Estimation}} & \textbf{Path Ranking}   \\ \cline{2-3} 
	& \multicolumn{1}{l|}{\textbf{MAE/MAPE/RMSE}}          & \textbf{MAE/$\tau$/$\rho$}    \\ \toprule[0.9pt]
	{w/o DVG}                         & \multicolumn{1}{c|}{4.66/27.49/7.06}                 & 0.13/0.79/0.83          \\ \hline
	{w/o VCM}                           & \multicolumn{1}{c|}{4.84/28.44/7.28}                 & 0.15/0.77/0.80          \\ \hline
	{w/o GCSM}                          & \multicolumn{1}{c|}{4.74/27.85/7.21}                 & 0.14/0.81/0/84          \\ \hline
	{DGCPath}                           & \multicolumn{1}{c|}{\textbf{4.66/27.49/7.06}}        & \textbf{0.11/0.82/0.86} \\ \toprule[1.5pt]
	\multirow{3}{*}{\textbf{Methods}} & \multicolumn{2}{c}{\textbf{Harbin}}                                           \\ \cline{2-3} 
	& \multicolumn{1}{c|}{\textbf{Travel Time Estimation}} & \textbf{Path Ranking}   \\ \cline{2-3} 
	& \multicolumn{1}{l|}{\textbf{MAE/MAPE/RMSE}}          & \textbf{MAE/$\tau$/$\rho$}    \\ \toprule[0.9pt]
	\emph{w/o DVG}                           & \multicolumn{1}{c|}{4.12/14.36/5.14}                 & 0.12/0.89/0.91          \\ \hline
	\emph{w/o VCM}                           & \multicolumn{1}{c|}{4.43/15.92/5.52}                 & 0.13/0.86/0.89          \\ \hline
	\emph{w/o GCSM}                          & \multicolumn{1}{c|}{4.24/14.94/5.29}                 & 0.13/0.88/0.90          \\ \hline
	\textsc{DGCPath}                           & \multicolumn{1}{c|}{\textbf{4.02/13.91/5.04}}                & \textbf{0.11/0.90/0.93}          \\ \bottomrule[1.5pt]
\end{tabular}
\label{tb:joint}
\caption{Effects of DVG, VCM Loss, and GCSM Loss}
\end{table}

\subsection{Path Recommendation}
To further verify the effectiveness and generality of \textsc{DGCPath}, we extend the experimental evaluation by incorporating the \textbf{Path Recommendation} task.

Following existing studies \cite{DBLP:journals/corr/abs-1907-04028}, the path taken by a user trajectory (e.g., path A) is labeled as 1, while alternative paths with the same origin and destination (e.g., paths B and C) are labeled as 0. The underlying intuition is that, given multiple candidate paths A, B, and C, the model should recommend the path actually chosen by the user (i.e., A).
We conduct experiments on the Aalborg and Chengdu datasets and evaluate performance using \textbf{Accuracy} (Acc., $\uparrow$), \textbf{Precision} (Pre., $\uparrow$), and \textbf{F1-score} ($\uparrow$), where higher values indicate better recommendation quality. Table~\ref{tab:PRP} reports the path recommendation results across all comparison methods. Overall, \textsc{DGCPath} consistently outperforms the unsupervised baselines, demonstrating its effectiveness in capturing user path preferences.

\subsection{Parameter Sensitivity Analysis}
We proceed to study three important hyper-parameter, including 1) temperature $t$, 2) diffusion step $T$, and embedding dimension $d$.

\noindent
\paragraph{Effect of Temperature $t$ of Distributional Contrastive Learning }To study the effect of the temperature $t$, we conduct a parameter study on Aalborg dataset, which is reported in Table~\ref{tab:temperature}. We observe that the performance of \textsc{DGCPath} varies with different temperatures. In particular, the best performance is achieved when $t=0.5$. This suggests that a moderately warm temperature effectively balances the sharpness of the contrastive objective, mitigating overly confident similarity assignments and leading to more stable optimization and improved representation quality.

\noindent
\paragraph{Effect of Embedding Dimensionality $d$ }To examine the impact of the path embedding dimensionality $d$, we conduct a parameter study on the Aalborg dataset. As reported in Table~\ref{tab:ed}, the performance of our model varies across different values of $d$. In particular, the best performance is achieved when $d=128$, indicating that this dimensionality provides a favorable balance between representation capacity and overfitting.

\begin{table}[t]
	\small
	\centering
	\caption{Evaluation on Path Recommendation Task.}
	{\renewcommand{\arraystretch}{1.3}
		\begin{tabular}{l|lll|lll}
			\toprule[1.5pt]
			\multirow{2}{*}{\textbf{Method}} & \multicolumn{3}{l|}{\textbf{Aalborg}}                                & \multicolumn{3}{l}{\textbf{Chengdu}}                                \\ \cline{2-7} 
			& \multicolumn{1}{l|}{\textbf{Acc.}} & \multicolumn{1}{l|}{\textbf{Pre.}} & \textbf{F1} & \multicolumn{1}{l|}{\textbf{Acc.}} & \multicolumn{1}{l|}{\textbf{Pre.}} & \textbf{F1}\\ \toprule[1pt]
			\textsc{Toast}                   & \multicolumn{1}{l|}{0.692}    & \multicolumn{1}{l|}{0.711}     &0.738      & \multicolumn{1}{l|}{0.774}    & \multicolumn{1}{l|}{0.785}     &0.806      \\ \hline
			\textsc{PIM}                     & \multicolumn{1}{l|}{0.722}    & \multicolumn{1}{l|}{0.739}     &0.762      & \multicolumn{1}{l|}{0.785}    & \multicolumn{1}{l|}{0.799}     &0.813      \\ \hline
			\textsc{START}                   & \multicolumn{1}{l|}{0.751}    & \multicolumn{1}{l|}{0.770}     &0.787      & \multicolumn{1}{l|}{0.804}    & \multicolumn{1}{l|}{0.826}     &0.853      \\ \hline
			\textsc{LightPath}               & \multicolumn{1}{l|}{0.752}    & \multicolumn{1}{l|}{0.768}     &0.790      & \multicolumn{1}{l|}{0.817}    & \multicolumn{1}{l|}{0.832}     &0.849     \\ \hline
			\textsc{RED }                    & \multicolumn{1}{l|}{0.785}    & \multicolumn{1}{l|}{0.811}     &0.830      & \multicolumn{1}{l|}{0.844}    & \multicolumn{1}{l|}{0.860}     & 0.879     \\ \hline
			
			\textsc{DGCPath}                 & \multicolumn{1}{l|}{\textbf{0.790}}    & \multicolumn{1}{l|}{\textbf{0.820}}     &\textbf{0.838}      & \multicolumn{1}{l|}{\textbf{0.851}}    & \multicolumn{1}{l|}{\textbf{0.865}}     &\textbf{0.883}      \\ \toprule[1.5pt]
	\end{tabular}}
	\label{tab:PRP}
\end{table}
\begin{table}[t]
	\caption{Effect of Temperature $t$ in Distributional Contrastive Learning }
	{\renewcommand{\arraystretch}{1.3}
		\begin{tabular}{l|llllll}
			\toprule[1.5pt]
			\multirow{3}{*}{\textbf{$t$}}                                                  & \multicolumn{6}{l}{\textbf{Aalborg}}                                                                                                                                                              \\ \cline{2-7} 
			& \multicolumn{3}{l|}{\textbf{Travel Time Estimation}}                                                        & \multicolumn{3}{l}{\textbf{Path Ranking}}                                           \\ \cline{2-7} 
			& \multicolumn{1}{l|}{\textbf{MAE}} & \multicolumn{1}{l|}{\textbf{MAPE}} & \multicolumn{1}{l|}{\textbf{RMSE}} & \multicolumn{1}{l|}{\textbf{MAE}} & \multicolumn{1}{l|}{\textbf{$\tau$}} & \textbf{$\rho$} \\ \toprule[1pt]
			0.1                                                 & \multicolumn{1}{l|}{2.194}        & \multicolumn{1}{l|}{13.758}          & \multicolumn{1}{l|}{3.762}         & \multicolumn{1}{l|}{0.107}         & \multicolumn{1}{l|}{0.839}         &0.851         \\ \hline
			0.5                                                & \multicolumn{1}{l|}{\textbf{2.131}}        & \multicolumn{1}{l|}{\textbf{13.654}}          & \multicolumn{1}{l|}{\textbf{3.725}}         & \multicolumn{1}{l|}{\textbf{0.107}}         & \multicolumn{1}{l|}{\textbf{0.846}}         & \textbf{0.870}         \\ \hline
			5                                                 & \multicolumn{1}{l|}{2.282}        & \multicolumn{1}{l|}{13.802}          & \multicolumn{1}{l|}{3.799}         & \multicolumn{1}{l|}{0.108}         & \multicolumn{1}{l|}{0.827}         &0.840          \\ \hline
			\toprule[1.5pt]
	\end{tabular}}
	\label{tab:temperature}
\end{table}
\begin{table}[htp]
	\caption{Effect of embedding dimensionality $d$ }
	{\renewcommand{\arraystretch}{1.3}
		\begin{tabular}{l|llllll}
			\toprule[1.5pt]
			\multirow{3}{*}{\textbf{$d$}}                                                  & \multicolumn{6}{l}{\textbf{Aalborg}}                                                                                                                                                              \\ \cline{2-7} 
			& \multicolumn{3}{l|}{\textbf{Travel Time Estimation}}                                                        & \multicolumn{3}{l}{\textbf{Path Ranking}}                                           \\ \cline{2-7} 
			& \multicolumn{1}{l|}{\textbf{MAE}} & \multicolumn{1}{l|}{\textbf{MAPE}} & \multicolumn{1}{l|}{\textbf{RMSE}} & \multicolumn{1}{l|}{\textbf{MAE}} & \multicolumn{1}{l|}{\textbf{$\tau$}} & \textbf{$\rho$} \\ \toprule[1pt]
			64                                                & \multicolumn{1}{l|}{2.234}        & \multicolumn{1}{l|}{13.826}          & \multicolumn{1}{l|}{3.922}         & \multicolumn{1}{l|}{0.109}         & \multicolumn{1}{l|}{0.833}         &0.851         \\ \hline
			128                                                & \multicolumn{1}{l|}{\textbf{2.131}}        & \multicolumn{1}{l|}{\textbf{13.654}}          & \multicolumn{1}{l|}{\textbf{3.725}}         & \multicolumn{1}{l|}{\textbf{0.107}}         & \multicolumn{1}{l|}{\textbf{0.846}}         & \textbf{0.870}         \\ \hline
			256                                                & \multicolumn{1}{l|}{2.212}        & \multicolumn{1}{l|}{13.799}          & \multicolumn{1}{l|}{3.879}         & \multicolumn{1}{l|}{0.107}         & \multicolumn{1}{l|}{0.838}         &0.855          \\ \hline
			\toprule[1.5pt]
	\end{tabular}}
	\label{tab:ed}
\end{table}
\begin{table}[t]
	\small
	\caption{Effect of diffusion step $T$ }
	{\renewcommand{\arraystretch}{1.3}
		\begin{tabular}{l|llllll}
			\toprule[1.5pt]
			\multirow{3}{*}{\textbf{$T$}}                                                  & \multicolumn{6}{l}{\textbf{Aalborg}}                                                                                                                                                              \\ \cline{2-7} 
			& \multicolumn{3}{l|}{\textbf{Travel Time Estimation}}                                                        & \multicolumn{3}{l}{\textbf{Path Ranking}}                                           \\ \cline{2-7} 
			& \multicolumn{1}{l|}{\textbf{MAE}} & \multicolumn{1}{l|}{\textbf{MAPE}} & \multicolumn{1}{l|}{\textbf{RMSE}} & \multicolumn{1}{l|}{\textbf{MAE}} & \multicolumn{1}{l|}{\textbf{$\tau$}} & \textbf{$\rho$} \\ \toprule[1pt]
			500                                                & \multicolumn{1}{l|}{2.653}        & \multicolumn{1}{l|}{14.021}          & \multicolumn{1}{l|}{4.103}         & \multicolumn{1}{l|}{0.110}         & \multicolumn{1}{l|}{0.820}         &0.843         \\ \hline
			1000                                                & \multicolumn{1}{l|}{\textbf{2.131}}        & \multicolumn{1}{l|}{\textbf{13.654}}          & \multicolumn{1}{l|}{\textbf{3.725}}         & \multicolumn{1}{l|}{\textbf{0.107}}         & \multicolumn{1}{l|}{\textbf{0.846}}         & \textbf{0.870}         \\ \hline
			1500                                                & \multicolumn{1}{l|}{2.354}        & \multicolumn{1}{l|}{13.873}          & \multicolumn{1}{l|}{3.913}         & \multicolumn{1}{l|}{0.108}         & \multicolumn{1}{l|}{0.837}         &0.858          \\ \hline
			\toprule[1.5pt]
	\end{tabular}}
	\label{tab:diffusion}
\end{table}

\noindent
\paragraph{Effect of Diffusion Step $T$ }To investigate the effect of the diffusion step $T$ of the view generator, we conduct a parameter study on the Aalborg dataset. The results are reported in Table~\ref{tab:diffusion}. We observe that \textsc{DGCPath} achieves the best performance when $T=1000$, indicating that an appropriate diffusion depth is crucial for generating informative yet diverse views. A smaller $T$ may lead to insufficient perturbation, while an excessively large $T$ can introduce unnecessary noise, both of which degrade representation quality.

\subsection{Effect of Dynamic Traffic State}
To verify the impact of dynamic traffic states, we conduct an additional experiment that incorporates temporal traffic information. Specifically, traffic flow sequences are constructed based on different time slots to capture time-varying traffic patterns. We denote \textsc{DGCPath} \emph{w/ DTS} as the variant of \textsc{DGCPath} that explicitly considers dynamic traffic states. The results in Table~\ref{tab:tt} show that incorporating dynamic traffic states further improves the performance of \textsc{DGCPath}.

\begin{table}[]
	\small
	\centering
	\caption{Effect of Dynamic Traffic State (DTS). }
	{\renewcommand{\arraystretch}{1.3}
		\begin{tabular}{l|ll}
			\toprule[1.5pt]
			\multirow{3}{*}{} & \multicolumn{2}{l}{\textbf{Aalborg}}                                        \\ \cline{2-3} 
			& \multicolumn{1}{l|}{\textbf{Travel Time Estimation}} & \textbf{Path Ranking} \\ \cline{2-3} 
			& \multicolumn{1}{l|}{\textbf{MAE/MAPE/RMSE}}          & \textbf{MAE/$\tau$/$\rho$}  \\ \hline
			\textsc{DGCPath}           & \multicolumn{1}{l|}{2.131/13.654/3.725}              & 0.107/0.846/0.870     \\ \hline
			\emph{w/ DTS}    & \multicolumn{1}{l|}{1.962/12.891/3.688}              & 0.105/0.856/0.879     \\ \toprule[1.5pt]
	\end{tabular}}
	\label{tab:tt}
\end{table}
\subsection{Model Scalability}
In this section, we investigate the scalability of \textsc{DGCPath} with respect to the size of the training data using the Aalborg dataset. Table~\ref{tab:MS} reports the results, where 20\%, 40\%, 60\%, 80\%, and 100\% denote the proportions of the training set used for model training. The results show a consistent performance improvement as the training data size increases, indicating that \textsc{DGCPath} effectively leverages larger datasets to enhance its modeling capability.

\begin{table}[htp]
	\small
	\caption{Model Scalalability Analysis }
	{\renewcommand{\arraystretch}{1.3}
		\begin{tabular}{l|llllll}
			\toprule[1.5pt]
			\multirow{3}{*}{}                                                  & \multicolumn{6}{l}{\textbf{Aalborg}}                                                                                                                                                              \\ \cline{2-7} 
			& \multicolumn{3}{l|}{\textbf{Travel Time Estimation}}                                                        & \multicolumn{3}{l}{\textbf{Path Ranking}}                                           \\ \cline{2-7} 
			& \multicolumn{1}{l|}{\textbf{MAE}} & \multicolumn{1}{l|}{\textbf{MAPE}} & \multicolumn{1}{l|}{\textbf{RMSE}} & \multicolumn{1}{l|}{\textbf{MAE}} & \multicolumn{1}{l|}{\textbf{$\tau$}} & \textbf{$\rho$} \\ \toprule[1pt]
			20\%                                                 & \multicolumn{1}{l|}{2.541}        & \multicolumn{1}{l|}{14.537}          & \multicolumn{1}{l|}{4.312}         & \multicolumn{1}{l|}{0.125}         & \multicolumn{1}{l|}{0.793}         &0.811         \\ \hline
			
			40\%                                                 & \multicolumn{1}{l|}{2.476}        & \multicolumn{1}{l|}{14.258}          & \multicolumn{1}{l|}{4.005}         & \multicolumn{1}{l|}{0.113}         & \multicolumn{1}{l|}{0.821}         &0.835        \\ \hline
			
			60\%                                                & \multicolumn{1}{l|}{2.334}        & \multicolumn{1}{l|}{13.902}          & \multicolumn{1}{l|}{3.922}         & \multicolumn{1}{l|}{0.109}         & \multicolumn{1}{l|}{0.836}         &0.847          \\ \hline
			
			80\%                                                 & \multicolumn{1}{l|}{2.229}        & \multicolumn{1}{l|}{13.811}          & \multicolumn{1}{l|}{3.823}         & \multicolumn{1}{l|}{0.108}         & \multicolumn{1}{l|}{0.840}         &0.859       \\ \hline
			
			100\%                                                & \multicolumn{1}{l|}{\textbf{2.131}}        & \multicolumn{1}{l|}{\textbf{13.654}}          & \multicolumn{1}{l|}{\textbf{3.725}}         & \multicolumn{1}{l|}{\textbf{0.107}}         & \multicolumn{1}{l|}{\textbf{0.846}}         & \textbf{0.870}         \\ 
			\toprule[1.5pt]
	\end{tabular}}
	\label{tab:MS}
\end{table}
\subsection{Robustness Analysis}
To approximate real-world scenarios with noisy trajectories, we injected Gaussian perturbations (e.g., $\sigma=0.03$) into each path representation and subsequently trained a regressor on the Aalborg and Chengdu datasets. The results, reported in Table~\ref{tab:RA}, demonstrate that \textsc{DGCPath} exhibits strong robustness to noisy representations and consistently outperforms all comparison methods, as it has a smaller performance drop percentage (\textbf{PDP}). For example, the \textbf{PDP} of \textsc{DGCPath} for travel time estimation (MAE) on the Aalborg dataset is 2.9\%, while the \textbf{PDP} of \textsc{RED} for travel time estimation on the Aalborg dataset is 2.9\%. In comparison, the \textbf{PDP} of \textsc{DGCPath} for path ranking ($\tau$) on the Aalborg dataset is 0.5\%, whereas the \textbf{PDP} of \textsc{RED} for path ranking ($\tau$) on the Aalborg dataset is 3.5\%. 

\begin{table}[]
	\small
	\centering
	\caption{Evaluation on Travel Time Estimation (TTE) and Path Ranking (PR) under Noisy Representations.}
	{\renewcommand{\arraystretch}{1.3}
		\begin{tabular}{l|ll}
			\toprule[1.5pt]
			\multirow{3}{*}{} & \multicolumn{2}{l}{\textbf{Aalborg}} \\ \cline{2-3} 
			& \multicolumn{1}{l|}{\textbf{TTE}} & \textbf{PR} \\ \cline{2-3} 
			& \multicolumn{1}{l|}{\textbf{MAE/MAPE/RMSE}} & \textbf{MAE/$\tau$/$\rho$} \\ \hline
			
			\textsc{Toast} & \multicolumn{1}{l|}{3.439/21.311/5.555} & 0.138/0.768/0.779 \\ \hline
			\textsc{Toast} \emph{w/ $\sigma$} & \multicolumn{1}{l|}{4.426/21.906/6.066} & 0.207/0.672/0.683 \\ \hline
			
			\textsc{PIM} & \multicolumn{1}{l|}{3.471/21.314/5.640} & 0.141/0.764/0.777 \\ \hline
			\textsc{PIM} \emph{w/ $\sigma$} & \multicolumn{1}{l|}{4.003/21.904/5.982} & 0.192/0.727/0.738 \\ \hline
			
			\textsc{START} & \multicolumn{1}{l|}{2.965/18.432/4.979} & 0.178/0.839/0.859 \\ \hline
			\textsc{START} \emph{w/ $\sigma$}& \multicolumn{1}{l|}{3.441/18.901/5.551} & 0.225/0.795/0.803 \\ \hline
			
			\textsc{LightPath} & \multicolumn{1}{l|}{2.920/18.254/4.932} & 0.118/0.781/0.791 \\ \hline
			\textsc{LightPath}  \emph{w/ $\sigma$} & \multicolumn{1}{l|}{3.232/18.673/5.412} & 0.156/0.736/0.744 \\ \hline
			
			\textsc{RED} & \multicolumn{1}{l|}{2.457/14.907/3.809} & 0.109/0.819/0.823 \\ \hline
			\textsc{RED} \emph{w/ $\sigma$} & \multicolumn{1}{l|}{2.527/14.999/3.903} & 0.127/0.790/0.805 \\ \hline
			
			\textsc{DGCPath} & \multicolumn{1}{l|}{2.131/13.654/3.725} & 0.107/0.846/0.870 \\ \hline
			\textsc{DGCPath} \emph{w/ $\sigma$} & \multicolumn{1}{l|}{2.193/13.924/3.873} & 0.109/0.842/0.862 \\ 
			\toprule[1.5pt]
		\end{tabular}
	}
	\label{tab:RA}
\end{table}

\section{Conclusion}
We propose \textsc{DGCPath}, a distribution-aware generative contrastive framework for path representation learning. Rather than relying on hand-crafted augmentations, \textsc{DGCPath} employs a learnable diffusion-based generator to produce two path views. It then uses a distribution-aware contrastive objective to align positive-pair distributions while separating positive and negative pairs. A generative cross-supervision module further enhances encoder training, yielding more generalizable path representations and improved performance across downstream tasks.

\bibliographystyle{named}
\bibliography{ijcai26}

\end{document}